\documentclass[preprint,12pt,authoryear]{elsarticle}

\usepackage{amssymb}

\journal{Expert Systems with Applications}

\begin{document}

\begin{frontmatter}



\title{Natural language processing on customer note data}

\author[label1]{Andrew Hilditch}
\ead{andrew.hilditch@mmu.ac.uk}
\author[label2]{David Webb}
\ead{dave.webb@autotrader.co.uk}
\author[label2]{Jozef Ba\v{c}a}
\ead{jozef.baca@autotrader.co.uk}
\author[label2]{Tom Armitage}
\ead{tom.armitage@autotrader.co.uk}
\author[label1]{Matthew Shardlow}
\ead{m.shardlow@mmu.ac.uk}
\author[label2]{Peter Appleby}
\ead{peter.appleby@autotrader.co.uk}
\affiliation[label1]{organization={Manchester Metropolitan University},
             city={Manchester},
             postcode={M15 6BH},
             country={UK}}
\affiliation[label2]{organization={Auto Trader Group},
             addressline={1 Tony Wilson Place},
             city={Manchester},
             postcode={M15 4FN},
             country={UK}}

\begin{abstract}
Automatic analysis of customer data for businesses is an area that is of interest to companies. Business to business data is studied rarely in academia due to the sensitive nature of such information. Applying natural language processing can speed up the analysis of prohibitively large sets of data. This paper addresses this subject and applies sentiment analysis, topic modelling and keyword extraction to a B2B data set. We show that accurate sentiment can be extracted from the notes automatically and the notes can be sorted by relevance into different topics. We see that without clear separation topics can lack relevance to a business context.
\end{abstract}




\end{frontmatter}


\section{Introduction}
\label{sec1}

To foster customer loyalty and provide assistance, companies frequently communicates with and collect feedback from their customers. Feedback is gathered in form of notes, that call handlers make after a conversation with a customer. Multiple studies have shown the impact of efficiently assessing customer feedback and implementing change, increasing customer satisfaction and loyalty \citep{Lam2013influence}; \citep{Azzam2014impact} ; \citep{Mulyono2018loyalty} as well as company revenues \citep{Phan2010model}.  
This study focusses on the customer notes collected by AutoTrader, a UK company that provides an online marketplace for UK car dealers.
Currently,  analysis is done manually on a subset of the notes with the intent to extract information that would be valuable for the company. 

These insights would be provided with Natural Language Processing (NLP). NLP refers to the branch of computer science concerned with giving computers the ability to understand text and spoken words in much the same way human beings can \citep{NLP}. The areas of interest were sentiment analysis, topic modelling and keyword extraction. 

\section{Related work}
The field of note analysis with NLP is well researched, with work being done on medical notes \citep{sheikhalishahi2019natural}, \citep{juhn2020artificial} and on data from social media networks like Twitter \citep{sanders2021unmasking} and Reddit \citep{okon2020natural}. Note analysis data is typically available in large quantities but requires pre-cleaning to remove irrelevant entries. It tends to include more errors than formal text and is typically not examined in great detail by hand. Work has been done to analyse the notes of those attempting suicide \citep{Pestian2010suicide}. This work classified the notes to attempt to predict repeated suicide attempts.
\subsection{Industrial applications of NLP}
Work has been done to look at NLP applications in industry. \citep{Kalyanatha2019advances} looked at different applications in areas such as finance and retail. Many of the current applications focus on chat bots \citep{Khan2021artificial}, \citep{Sari2020chatbot}. They look at banking and social media networks. Work has also been done to examine the use of NLP in the construction sector \citep{Ding2022applications}. This worked examined the use of NLP for risk management and building information modelling among other uses.

\subsection{Sentiment analysis}
Sentiment analysis is a subject of huge importance in data science that has gained significantly in prevalence over the last decade or so, with more than 99\% of papers published on the subject coming after 2004 due to the vast expansion of unstructured text based datasets available \citep{Mantyla2018evolution}. Sentiment analysis can be performed using both supervised and unsupervised methodologies, we will focus on the unsupervised approach.

Many papers that study sentiment using NLP have been published. One is the aforementioned study looking at Twitter sentiment \citep{okon2020natural}. There are many other papers looking at Twitter data \citep{Kanakaraj2015performance}, \citep{Hasan2019sentiment}. Previous researchers in the area have used lexicon approaches to study sentiment for topics within a document \citep{Nasukawa2003sentiment}. The work on sentiment in this paper is in section three.

\subsection{Transformers}
NLP has changed since the introduction of the Transformer architecture \citep{Vaswani2017attention}. It led to a number of papers that utilised and developed the architecture such as  Bidirectional Encoder Representations from Transformers (BERT) \citep{Devlin2018bert} and RoBERTa \citep{Liu2019roberta}.  Like a neural network, Transformers work by utilising a deep learning model where all output elements are connected to all input elements, with the weightings between them calculated dynamically based on the connection. BERT differs however in that the text is read by the algorithm both forwards and backwards at the same time, allowing text to attenuate with itself rather than an output layer. This feature drastically increases the speed at which BERT can train and allows entire sentences to be used as inputs rather than tokenized data, meaning BERT is capable of contextualising words within their sentence structure. This last point has proved revolutionary in the world of NLP research as prior to Transformers all works had to be performed with word tokens taken in isolation, severely limiting their use to the specific data they were trained upon.

BERT was made open source by Google in 2018 having been trained on the entire content of the English language version of Wikipedia \citep{Devlin2018bert}. Since then BERT has been converted into libraries capable of a wide range of NLP tasks such as sentiment analysis, sentence completion and text summarisation. Additionally, these libraries come with added advantage of being able to fine tune the algorithm with relatively small datasets to increase the specificity to a corpus of choice (such as the AutoTrader note data), although classification tasks such as sentiment analysis will require labelled data. Utilisation of these libraries is easy to implement but can be computationally expensive, often requiring access to a GPU to run in efficient time frames which should also be considered before use. Many of these new Transformer models can now be found on the HuggingFace \citep{Wolf2019hugging} website where the Transformers library is located \citep{Brasoveanu2020visualizing}. The models used in this paper were obtained from the HuggingFace library \citep{Hugging2023AI}.

\subsection{Topic modelling}
Topic modelling provides an unsupervised approach to topic allocation in texts, with a broad range of complexities. It will be explored in sections four and five of this paper Simple approaches such as Latent Semantic Indexing (LSI) \citep{Hofmann1999probabilistic} involve the vectorisation of texts within a corpus and grouping together based on co-sine similarity (an effective measure used to compare similarity of vectors \citep{Han2001data}). Such methods are quick to implement and require little resources but come with the large disadvantage that the nature of the topic groupings remains unknown, making the results very difficult to interpret. More sophisticated methods tend to be based on more complex algorithms such as neural networks, such as the work from \citep{Gavval2019cuda} which explores the use of self-organising maps (SOMs) to reduce dimensionality within the data and create an interpretable 2D map of topics. But whilst having the advantage of interpretability these methods are often computationally expensive and can tie up valuable resources within an organisation.

Latent Dirichlet Allocation (LDA) \citep{Blei2003latent} provides a middle ground between overly simply non-interpretable and overly-complex resource heavy topic modelling techniques, and is one of the most commonly used methods in the field \citep{Jelodar2019latent}. LDA involves creating a latent layer of topics within a dataset where words that are likely to be found near each other within texts are grouped. Each text within a corpus is then evaluated for a percentage match with each of the topics in the latent layer to allow allocation. One of the drawbacks with LDA is that as a statistical approach, interpretation of the topics is still required to achieve sensible results, just because words are statistically found near each other does not necessarily mean they will be considered related by a human observer. This means LDA topic modelling can require extensive hyperparameter tuning to produce good results but having a sector expert on the texts at hand to perform this can add a built-in sanity check step to the evaluation. This is one of the methods used in section four of this paper.

\subsection{Keyword extraction}
Keyword extraction (KE) has been used to improve the efficiency of other NLP methods, most notably Information Retrieval \citep{Yang2019toward}. These methods then don’t have to processes the whole text of a document, but only text that carries the subject of the document. Use with other NLP applications implies that the KE algorithm should be quick, efficient, robust and easily applied to new domains. This led to a new method being created called Rapid Automatic Keyword Extraction (RAKE) \citep{Rose2010automatic}. RAKE is an unsupervised, language-independent document-oriented method for extracting keywords from individual documents. RAKE’s creators observed that keywords often appear as a combination of multiple words rather than a single word, and almost never contain any stop words or punctuation. RAKE considers these non-stop words as the main candidates then uses a graph-based approach to capture the co-occurrences of these candidates, which are used to calculate a score to choose the best keywords.

Deep learning approaches of KE are not common, as Deep neural networks (DNNs) require large annotated datasets. The introduction of transformers has inspired scientists to propose new methods utilising the self-attention layers without the need of labelled datasets. In 2019 a new approach was proposed, called the Self-supervised Contextual Keyword and Keyphrase Retrieval with Self-labelling (SCKKRS) \citep{Sharma2019self}. The aim of SCKKRS is for it to be useful for both long as well as short text inputs to extract keywords and keyphrases. SCKKRS uses feature vectors to obtain a keyword that is most similar to the meaning of the sentence. It obtains the feature vectors using BERT.

KeyBERT is an implementation of SCKKRS approach, which combines the feature vectors containing the meanings from BERT with statistical n-gram approach. KeyBERT works by extracting groups of words, which have the highest similarity between their embeddings and the sentence embedding \citep{Rao2022keyword}. This approach will be used in section five of this paper.

\subsection{Clustering}
Generic K-Means algorithm, as described by in 2013 \citep{Kodinariya2013review}, is an unsupervised learning algorithm that does not have high computational complexity. It has been utilised in many fields of NLP. The process of K-Means provides an easy solution to a classification problem of classifying data into known number of clusters. The main downside of the K-Means algorithm is the need to know the number of clusters. There are several approaches to find the best k value, but they don’t generally produce an answer without running the algorithm several times with different values, costing time and computational resources. K-Means clustering is commonly used in the NLP problem of Topic Modelling (TM). The method first applies a NLP function, which captures the textual content and then uses the K-Means algorithm to categorise data into clusters \citep{Curiskis2020evaluation}. In the same study \citep{Curiskis2020evaluation}, a combination of Doc2Vec feature representation and K-Means clustering gave the best performance across two datasets.

\section{Sentiment analysis}
\subsection{Introduction}
Sentiment analysis is an important feature of the note analysis for AutoTrader as it provides a measure on customer opinion towards product and policy changes implemented by the company. Current sentiment analysis works within AutoTrader are done manually by dedicated sector experts, but this method is problematic for three reasons.
\begin{enumerate}
    \item The process is labour intensive and requires full reading and comprehension of the notes from a sector expert, and with thousands of notes received a month processing every one is not realistic, meaning arbitrary prioritisation methods such as selection based on anecdotal evidence must be used.
    \item Individual sector experts may carry inherent bias in their assessment of note sentiment leading to added variation in the results.
    \item Manual sentiment analysis is very difficult to classify in a granular scale (i.e. a scored value rather than simply positive or negative), a feature which has been highlighted of importance by the AutoTrader team in allowing them to identify an underlying baseline sentiment for their customer feedback.

\end{enumerate}
These issues can all be solved by automated unsupervised sentiment analysis techniques which allow for efficient processing in a non-biased fashion with the option for continuous scoring depending on the approach used.

Automated unsupervised sentiment analysis approaches do come with an issue for the AutoTrader note data. Unsupervised techniques will still be trained on available datasets to the creator, which whilst not making them specific to that dataset will provide a better accuracy for similarly structured data. A literature search has found few examples of techniques that are tuned on B2B customer feedback data similar to the AutoTrader notes. We hypothesise that this is due to B2B customer feedback containing sensitive data which companies would be reluctant to disclose for academic publishing. Most unsupervised techniques are therefore trained on publicly available B2C data where customer feedback is collected from open access public sources such as Amazon or Yelp. B2B data differs from B2C (business to customer) data as it is typically collected using dyadic (person to person) rather than automated (person to device) techniques \citep{Murphy2018communication}. Dyadic conversation tends to more personable than automated, with a level of mediation which tones downs language used \citep{Aguilar2016dyadic}. This toning down of language may lead to a miscalibration of any scoring systems used to gauge sentiment within a technique that this trained using automated communication with more extreme language usage.
\subsection{Methodology}
\begin{figure}[h]
    \centering
    \includegraphics[width=1\textwidth]{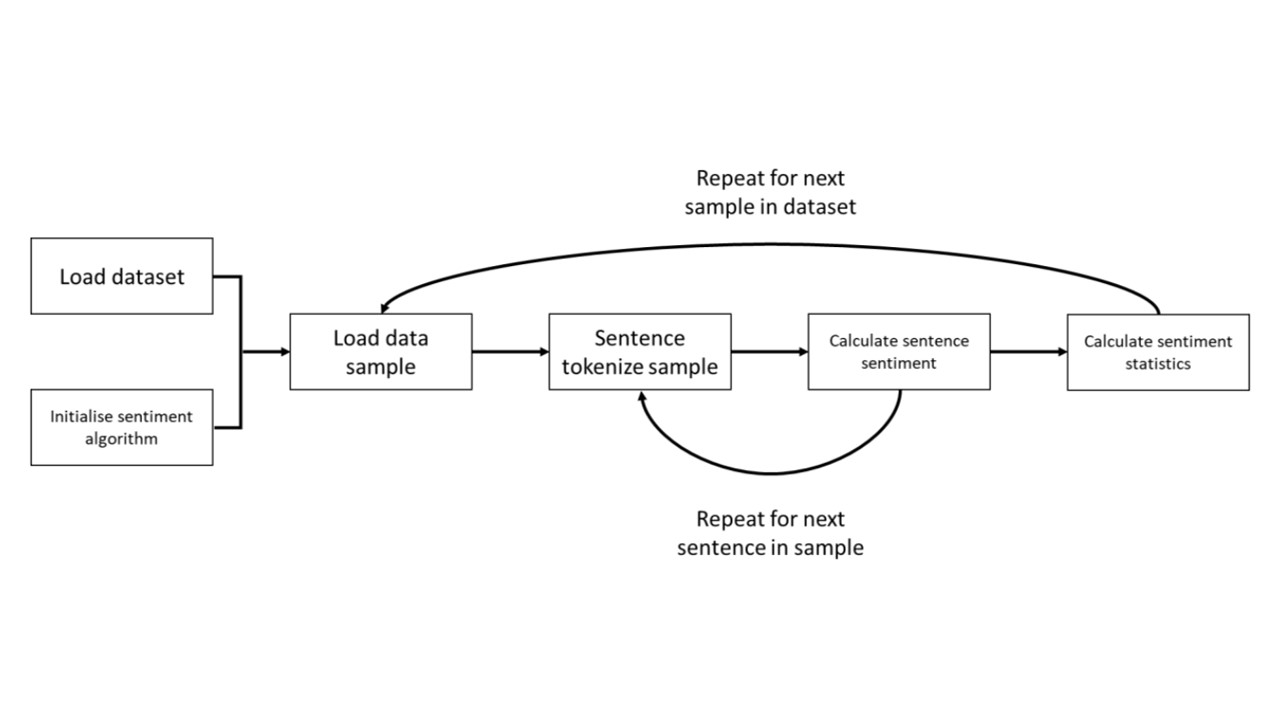}
    \caption{Flow diagram of the sentiment analysis process.}
    \label{fig:Sentiment-flow}
\end{figure}

Figure \ref{fig:Sentiment-flow} shows the method for calculating the sentiment of the data set.

\begin{enumerate}
    \item Load dataset
    \begin{enumerate}
        \item Dataset is loaded into the notebook as a pandas data frame
    \end{enumerate}
    \item Initialise sentiment algorithm
    \begin{enumerate}
        \item Sentiment algorithm of choice library is loading into the notebook and initialised based on instructions from the HuggingFace directions
        \item Note that no hyperparameter choices are required as the HuggingFace library does not support this
    \end{enumerate}
    \item Load data sample
    \begin{enumerate}
        \item Loop established to cycle through each text in the loaded dataset in turn through steps 4 to 6
    \end{enumerate}
    \item Sentence tokenize data
    \begin{enumerate}
        \item Individual text is tokenized at a sentence level, with a loop generated to run through each tokenized sentence in turn through step 5
        \item Sentences longer than 522 characters are truncated due to the word limit of the HuggingFace algorithm
        \item Note that sentence tokenization was preferred as it produced better results than analysing each text as a whole
    \end{enumerate}
    \item Calculate sentence sentiment
    \begin{enumerate}
        \item Sentence text is run through the sentiment analysis function to calculate the score
        \item HuggingFace outputs a score and a sentiment label (e.g. [{`label': `POSITIVE', `score': 0.9998}], score values were extracted from this output with negative labelled score multiplied by -1
    \end{enumerate}
    \item Data is collected and appended to the text entry data row in this initial loaded data frame, collected data includes
    \begin{enumerate}
        \item Average sentence sentiment
        \item Negative sentiment sentence count
        \item Max negative sentence score
        \item Positive sentence count
        \item Max positive sentence score
    \end{enumerate}
\end{enumerate}

\begin{figure}[h]
    \centering
    \includegraphics[width=0.8\textwidth]{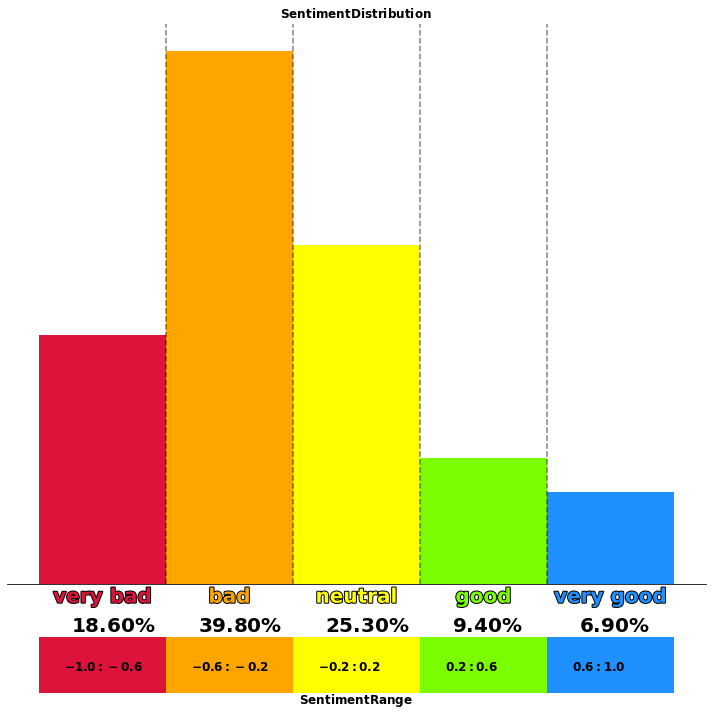}
    \caption{Bar chart showing distribution of annotated sentiment.}
    \label{fig:Distributed annotated sentiment}
\end{figure}

A subset of 1000 of the notes were annotated with sentiment sorted into five categories. These are very bad, bad, neutral, good, very good. The breakdown is shown in figure \ref{fig:Distributed annotated sentiment}.

\subsection{Results}
\begin{figure}[h]
    \centering
    \includegraphics[width=0.8\textwidth]{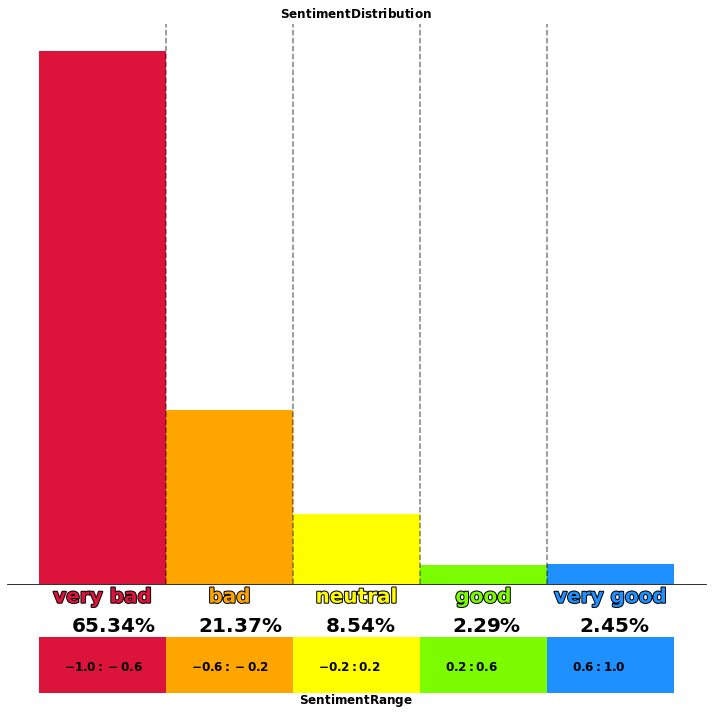}
    \caption{Bar chart showing distribution of calculated sentiment.}
    \label{fig:Distributed calculated sentiment}
\end{figure}

Initial analysis of the sentiment distribution contained within the AutoTrader note data showed that over 65\% of the notes were classified as extremely negative (figure \ref{fig:Distributed calculated sentiment}), and almost 87\% of all notes showed a negative sentiment score. The reason for this is that the notes worked on were marked as feedback, which tends towards negative sentiment. Additionally research has shown that sector experts tend to provide more negative feedback than novices \citep{Finkelstein2012tell} in order to help companies develop their product, and as AutoTrader specialises in B2B trading with sector experts it is not unexpected that their customer feedback also reflects this.

Also noted in the data are notes with non-negative sentiment. Neutral sentiment notes tended to be created by feedback which was purely informational with no information on customer mood shared, for example:
\begin{center}
  \emph{“he had his 3 lads running around his feet and his wife was doing the big shop asked i call back either this afternoon or monday.”}   
\end{center}

whilst on analysis the extreme positive notes appeared to be mostly single sentence with a focus on a single positive topic, for example:
\begin{center}
   \emph{“confirmed the communication with him he is happy with what we have mentioned”} 
\end{center}

Though not seen as statistically significant, it is noted that longer multiple sentence notes produced fewer extreme scores due to the average sentiment being taken from multiple
different topic sentences.

\begin{figure}[h]
    \centering
    \includegraphics[width=1\textwidth]{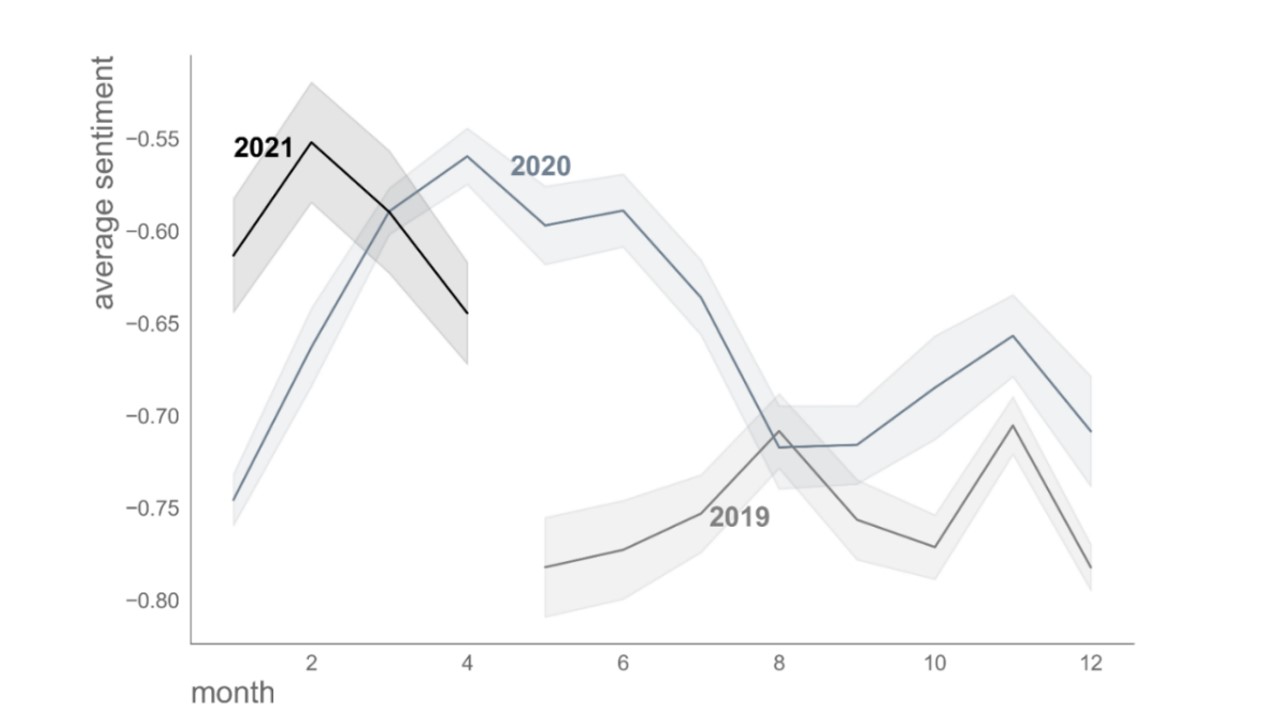}
    \caption{Chart displaying the change in sentiment over time.}
    \label{fig:Sentiment-over-time}
\end{figure}

Reviewing the change in note sentiment over time (figure \ref{fig:Sentiment-over-time}) also revealed some interesting trends in the data. We can observe spikes in the relative note sentiment during April – July 2020 and in January – February 2021, coinciding with the full lockdown dates for the UK. This increase in sentiment correlated with a popular policy from the company. Reflecting the overall sentiment distribution, the general baseline sentiment value over time is negative, between -0.80 and -0.65. This overall negativity is not an issue however as the changing sentiment over time was the focus of the research. The interest can be placed instead on the relative sentiment values with respect to the baseline when assessing for the outcomes of any further policy changes affecting customers.

The calculated sentiment when compared to the annotated sentiment was far more negative. Using a pairwise comparison between the two the model achieves an accuracy score of 24\% when compared against all five categories. When compared for three categories, (positive, negative and neutral) there is an agreement of 60\%.
\subsection{Recommendations}
There are a few suggestions that we have for anyone looking to perform similar analysis. 
\begin{enumerate}
    \item Look to reduce granularity wherever possible to speed up analysis. The analysis done here can take hours of compute time to calculate so the more operations that can be performed on the entire dataset the more time can be saved. Be aware that this cannot be done for some operations.
    \item Be cautious of the reserved language and reduced punctuation present in B2B data. This could see worse performance for models trained on B2C data when predicting sentiment. Our dataset is examining feedback so be aware and check with sector experts as to what to expect from your dataset.
    \item Look at different tokenisation levels such as sentences paragraphs or full documents. Higher level analysis will return neutral sentiment on a more frequent basis. When we examined sentiment at a whole note level the longer notes tended towards neutrality.
    \item Examine analogous labels to your subject area, this will better allow you to compare results. So the note data here is best compared to customer review data. It would be better compared to B2B reviews but these are not publicly accessible.
\end{enumerate}
\subsection{Conclusion}
In conclusion we have demonstrated a robust method for assessing the sentiment of the AutoTrader note library utilising a deep learning based approach with the HuggingFace library. Confirmation from the AutoTrader team that the general trends seen within the data fit with expected outcomes of policy changes implemented by the company also acted as a sanity check for the efficacy with their data.

It is noted that a large portion of the notes were seen to be extremely negative with the note set, which despite justification of the outcome does cause some concern. As the focus for the analysis is for change in sentiment however this may not prove an issue as relative sentiment is considered for drawing conclusions. The comparison to the annotated data shows that the implemented method has a tendency to misjudge the strength of the sentiment within a message. This could be due to the AutoTrader data containing unfamiliar terms. The other reason could be that the removal of personal identifiable data could have left sentences without subjects to attribute sentiment to. This model could be more robustly calculated with the use of more than one annotator to verify the reliability of the annotations. Transfer learning \citep{Weiss2016survey} could be used to allow the model to understand the data better.
\section{Topic modelling utilising Latent Dirichlet Allocation}
\subsection{Introduction}
If the notes could be sorted automatically into relevant categories it would be easier to analyse issues in different areas. Issues are raised anecdotally from team members collecting the notes and evidence is searched for within the data using sector expert derived keywords. This approach can lead to issues within identifying target areas for policy improvements, in particular “firefighting” where only big issues and complaints are dealt with as they grab the most attention, while smaller issues that may be more prevalent can remain untouched. Additionally, even issues that have been identified may be missed if customer feedback cannot be found due to the use of incorrect keywords during evidence searches. Approaches to tackle this issue have been undertaken such as adding selective hashtags to key words in customer note data and initiating topic check boxes but these have so far been unsuccessful.

Topic modelling can provide a different approach to the problem for AutoTrader. It utilises a data driven approach where the topics of interest are derived from the data itself rather than expert led anecdotal evidence which may contain unintended bias. Topic modelling however is not flawless as human interaction is still required to make sense of the topics suggested by the chosen methodology, but this may present itself as an additional opportunity. Human interaction with the algorithm will ensure investment from the interested parties and can be used to sense check the model during training rather than finding the results are faulty at the analysis stage. For this section we shall demonstrate the use of LDA topic modelling with the AutoTrader note data, with a view to demonstrate the types of note data that can be extracted and how the data can be merged with the sentiment analysis results from section three to be used for further analysis.
\subsection{Methodology}
\begin{figure}[h]
    \centering
    \includegraphics[width=1\textwidth]{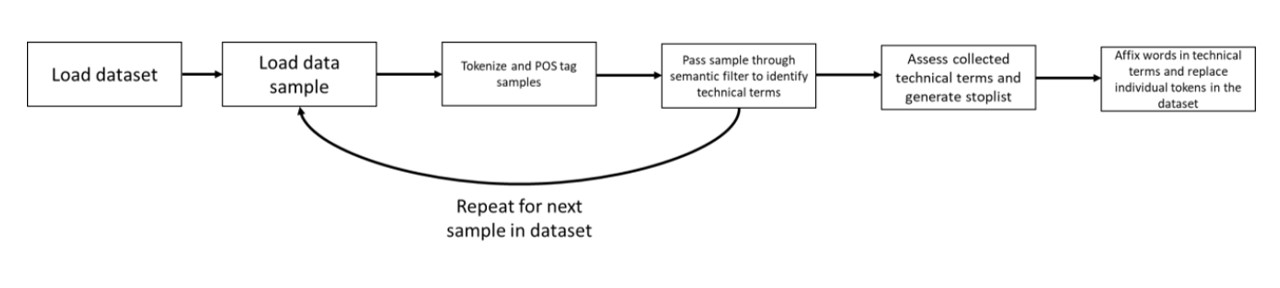}
    \caption{Flow diagram of the technical term extraction process.}
    \label{fig:Tech-term-flow}
\end{figure}

The data used here was the same as in section three with the additional sentiment analysis added to the data. Additional data preparation was performed via technical term extraction and joining into n-grams, using the following flow process outlined in figure \ref{fig:Tech-term-flow}, with a step-by-step breakdown detailed below.

\begin{enumerate}
    \item Load dataset
    \begin{enumerate}
        \item Dataset is loaded into the notebook as a pandas data frame
    \end{enumerate}
    \item Load data sample
    \begin{enumerate}
        \item Loop established to cycle through each text in the loaded dataset in turn through steps 3 and 4
    \end{enumerate}
    \item Tokenize and Part-of-Speech (POS) tag samples
    \begin{enumerate}
        \item Notes are tokenized at the word level
        \item Each token is POS tagged with a semantic label using the nltk POS tagging library \citep{Bird2019categorizing}
    \end{enumerate}
    \item Pass sample through semantic filter to identify n-grams
    \begin{enumerate}
        \item Samples are passed through the technical term semantic filter defined by \citet{Justedon1995technical}
        \item Technical terms limited to bi-grams and tri-grams as longer terms are typically only associated with highly technical fields \citep{Justedon1995technical}
    \end{enumerate}
    \item Assess collected technical terms and generate stoplist 
    \begin{enumerate}
        \item Top 100 most frequent terms manually assessed and common terminology phrases removed
        \item Top 100 most frequent terms reassessed after stoplist terms removed and additional identified stop terms removed
        \item Process repeated until top 100 phrases clear of common terminology phrases
        \item Process repeated until top 100 phrases clear of common terminology phrases - c-score weightings of the technical terms calculated using the method defined by \citet{Frantzi2000automatic}
    \end{enumerate}
    \item Affix technical term words in dataset and replace
    \begin{enumerate}
        \item All technical term tokens within the list adjoined with an underscore to create a single token (e.g. “new”, “cars” becomes “new\_cars”)
        \item Technical terms individual tokens removed from the dataset
    \end{enumerate}
\end{enumerate}

\begin{figure}[h]
    \centering
    \includegraphics[width=1\textwidth]{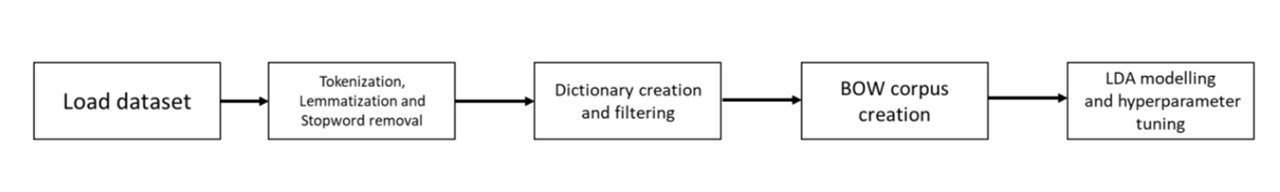}
    \caption{Flow diagram of the topic modelling process.}
    \label{fig:TM-flow}
\end{figure}

Figure \ref{fig:TM-flow} shows the process by which the topic modelling is done. Again the flow chart is explained in more detail below.

\begin{enumerate}
    \item Load dataset
    \begin{enumerate}
        \item Dataset is loaded into the notebook as a pandas data frame
    \end{enumerate}
    \item Tokenization, lemmatization and stopword removal
    \begin{enumerate}
        \item Each note in the corpus is run through in turn
        \item Tokenization is performed at the word level using the gensim.simple\_preprocess library in python \citep{Rehurek2022utils}
        \item Lemmatization of tokens is performed using the nltk WordNetLemmatizer library \citep{NLTK2023stem}
        \item Stopwords are removed from each note using the nltk library stopword corpus \citep{NLTK2023corpora}
    \end{enumerate}
    \item Dictionary creation and filtering
    \begin{enumerate}
        \item A dictionary is created from the entire pre-processed corpus using the gensim.corpora dictionary library \citep{Rehurek2022corpora}
        \item Once created the dictionary is filtered to remove noise from extreme case tokens
        \begin{enumerate}
            \item Tokens featured in 5 or less notes are removed
            \item Top 100,000 used terms are reserved
            \item Limit of the number of texts a token can feature in is also used in the dictionary filter which is tuned by the expert operator as a hyperparameter
        \end{enumerate}
    \end{enumerate}
    \item BOW corpus creation
    \begin{enumerate}
        \item Each note is compared against the dictionary to create a BOW note using the gensim.corpora dictionary library doc2bow feature \citep{Rehurek2022corpora} and tokens in the dictionary are retained and their frequency recorded
    \end{enumerate}
    \item LDA modelling and hyperparameter tuning
    \begin{enumerate}
        \item LDA model is generated using gensim.models.ldamulticore library \citep{Rehurek2022models} using the created dictionary and BOW corpus as inputs
        \item Hyperparameters tuned as follows
        \begin{enumerate}
            \item Number of passes set to ten (experimentation showed consistent convergence of the model at this number)
            \item Number of topics to be tuned by an expert user assessing the resultant topics for sense
            \item Minimum probability level for a note to be contained within a topic to be tuned by an expert user assessing the resultant topics for sense
        \end{enumerate}
    \end{enumerate}
\end{enumerate}
\subsection{Results}
\begin{figure}[h]
    \centering
    \includegraphics[width=1\textwidth]{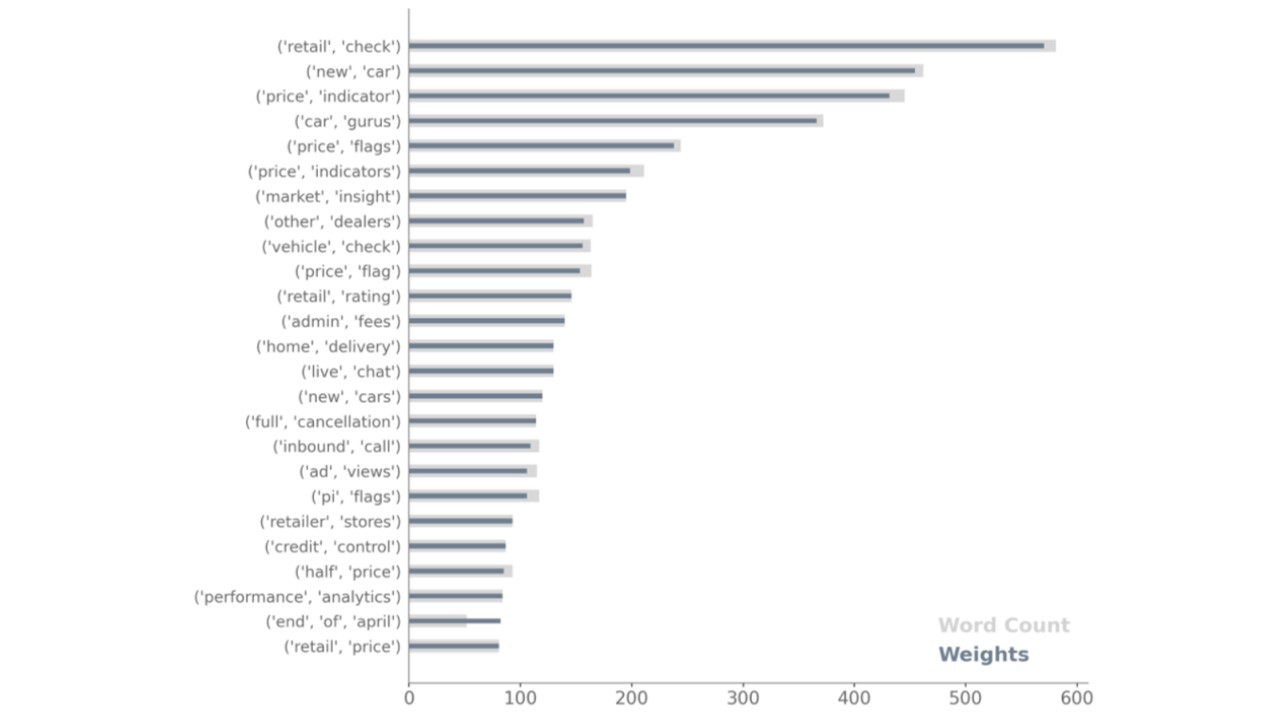}
    \caption{Identified technical term n-grams identified from the Autotrader note corpus. Weights denote the c-score given to each of the terms.}
    \label{fig:Tech-terms}
\end{figure}

A list of the top 25 technical term n-grams identified from the AutoTrader corpus is shown in Figure \ref{fig:Tech-terms}. Noted from the results is that bi-grams dominate the list, justifying the decision to only look for bi and tri-grams within the corpus. The only tri-gram within the list is “end-of-April”, which despite being a generic date based term holds weight as this is the typical annual date of a customer based policy change and is considered the end of the financial year. Also observed is that the word frequency and weight (c-score) drop off significantly after the first 4 n-grams on the list, indicating that the set of notes analysed are potentially dominated by topics revolving around those terms. In fact, the weighting for the third term on the list (“price-indicator”) appears to have been dampened as the same term is referred to in several different variations (“price-flag”, “price-indicators”, “price-flag”, “pi-flags”), which, if combined, would significantly increase the weighting of the term. This highlights a potential process improvement for data collection where terms with the same meaning referred to in different ways can be combined in the initial cleaning stage of note processing. If this is done before running through the sentiment and topic analysis algorithms the analysis would be better sorted by toipcs.

As previously described the topic modelling hyperparameter tuning was performed alongside a sector expert from AutoTrader to ensure sensible results. Table 1 shows the final satisfactory set of identified topics from the corpus. 10 topics were identified using a heavy dictionary filter with no terms featured in more than 5\% of the notes included and a minimum probability level of 0.005\% of being within a topic. Feedback from the AutoTrader team indicated that most of the topics identified showed consistent language and could be classified as relevant topics within the business (labelled as such in table \ref{table:1}), with only one of the topics being deemed as nonsense (labelled as “no recognised subject”). Also identified by the AutoTrader team were two topics not expected to be within the notes, namely “live chat” and “package”, which would not have searched for using the old knowledge led approach.

\begin{table}[!htbp]
\begin{tabular}{|p{0.88\textwidth}|p{0.15\textwidth}|}
\hline
High Probability Words & Suggested topic\\
\hline
0.018*"quote" + 0.013*"ebay" + 0.010*"finance" + 0.009*"premium" +
0.009*"level" + 0.008*"performance" + 0.007*"new\_car" +
0.007*"expensive" + 0.007*"struggle" + 0.006*"advance" & package\\
\hline
0.019*"data" + 0.015*"flag" + 0.013*"meet" + 0.011*"retail\_check" +
0.010*"price\_indicator" + 0.010*"spec" + 0.008*"sit" +
0.008*"valuations" + 0.007*"group" + 0.007*"price\_flags" & price indicator flags\\
\hline
0.012*"request" + 0.012*"admin\_fees" + 0.011*"video" + 0.007*"find" +
0.007*"image" + 0.006*"actually" + 0.006*"query" + 0.006*"frustrate" +
0.005*"spec" + 0.005*"unhappy" & unhappy\\
\hline
0.017*"image" + 0.013*"rat" + 0.012*"new\_car" + 0.010*"highly" + 0.009*"upload" + 0.008*"reply" + 0.008*"award" + 
0.007*"consumers" + 0.006*"info" + 0.006*"message" & live chat\\
\hline
0.012*"text" + 0.011*"valuations" + 0.011*"product" + 0.010*"chat" +
0.009*"lose" + 0.007*"tech" + 0.007*"margin" + 0.006*"platform" +
0.006*"retail" + 0.006*"higher" & valuations\\
\hline
0.010*"retract" + 0.008*"close" + 0.008*"open" + 0.007*"watch" +
0.007*"webinar" + 0.006*"book" + 0.006*"phone" + 0.006*"process" +
0.006*"answer" + 0.006*"charge" & process related\\
\hline
0.011*"staff" + 0.010*"coronavirus" + 0.010*"reduce" +
0.010*"lockdown" + 0.009*"canx" + 0.009*"plan" + 0.008*"struggle" +
0.008*"online" + 0.008*"june" + 0.008*"continue" & coronavirus\\
\hline
0.016*"lockdown" + 0.010*"june" + 0.010*"collect" + 0.008*"open" +
0.008*"retract" + 0.007*"confuse" + 0.007*"follow" + 0.007*"aware" +
0.007*"extend" + 0.007*"appreciate" & lockdown extensions\\
\hline
0.061*"xxxemailxxx" + 0.023*"subject" + 0.015*"group" + 0.013*"kind"
+ 0.009*"xxxtelephonexxx" + 0.008*"sit" + 0.008*"retail" + 0.007*"lead"
+ 0.007*"manheim" + 0.006*"option" & no recognised subject\\
\hline
0.027*"year" + 0.016*"experian" + 0.008*"car\_gurus" + 0.007*"ebay" +
0.006*"meet" + 0.006*"zuto" + 0.006*"july" + 0.005*"achieve" +
0.005*"award" + 0.005*"normal" & rival valuation products\\
\hline
\end{tabular}
\caption{Topic modelling results from the AutoTrader note corpus, with sector expert led topic naming suggestions.}
\label{table:1}
\end{table}

\begin{figure}[h]
    \centering
    \includegraphics[width=1\textwidth]{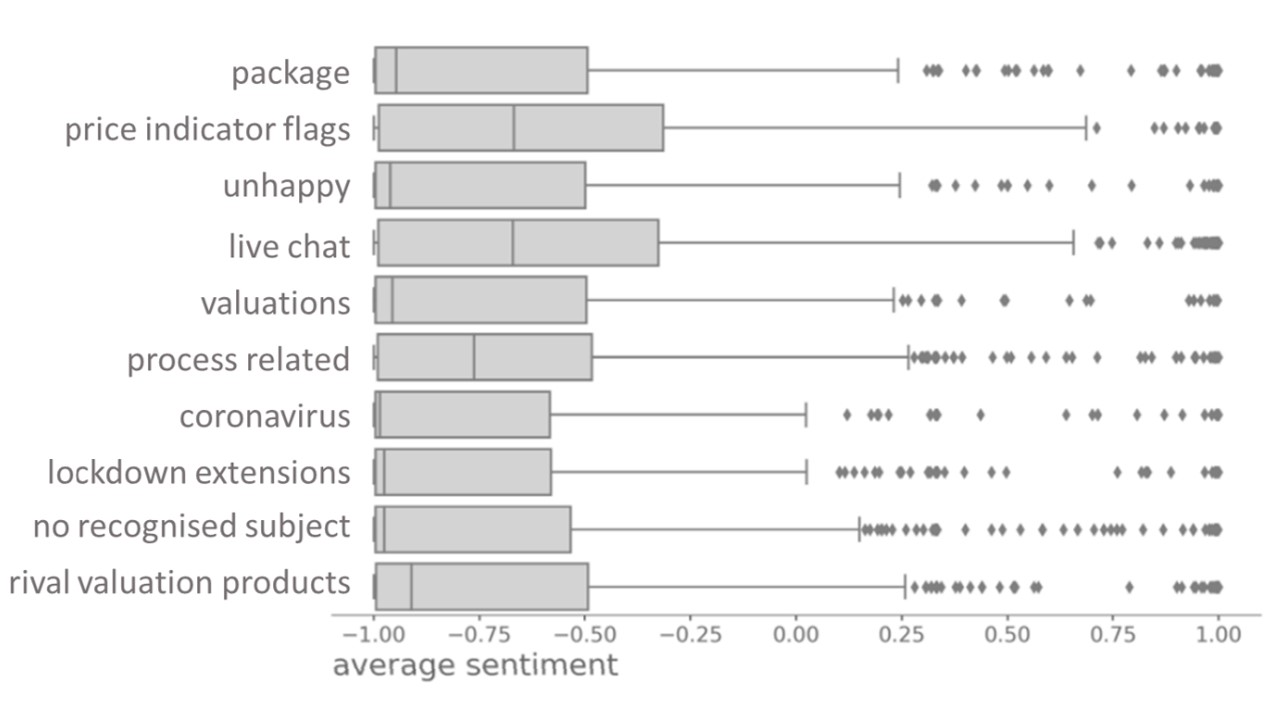}
    \caption{Mean average note sentiment distribution for the unidentified topics within the AutoTrader dataset.}
    \label{fig:Mean-flow}
\end{figure}

In addition to assessing the topic modelling in isolation the results were combined with the earlier sentiment analysis works to assess the variation in sentiment amongst the identified topics. Figure \ref{fig:Mean-flow} shows the distributions of the mean average note sentiment over each of the identified topics within the note corpus. Analysis shows that each of the topics follows the general trend of the overall note sentiment identified in section 3.3, with the notes showing a general negative sentiment skew with a long tail and outliers present for extreme positive sentiment. From the chart we can see that the “price indicator flags” and “live chat” topics show a higher median sentiment value and wider distribution towards the positive than the other identified topic. These significant differences would indicate that these topics warrant further investigation to assess why their average sentiment differs from the underlying population. Their higher average sentiments are due to the same positive sentiment spike as seen in section three and correlate with the first coronavirus lockdown.

\subsection{Recommendations}
There are also suggestions that we would have when examining topic modelling for B2B data.
\begin{enumerate}
    \item The first is that where possible use the bespoke knowledge of sector experts. They will likely have a great understanding of acronyms and keywords. They will also understand the context behind topics and how relevant they are to the area of study.
    \item Beware possible bias from experts on certain topics, let the data guide the overall analysis. The experts may already have decided their view on a topic or have a vested interest in the sentiment analysis of a topic. The underlying data will inform as to whether that view is justifiable or not.
    \item Use harsh dictionary filters to remove terms that are too frequent. These terms can distort the desired keyword output. This also removes a greater amount of stop words.
    \item Apply advanced lemmatisation for the relevant area as well as technical N-gram creation which will be required to examine specific sector terms. This will allow for the adaptation to the terminology and technical language found within the dataset that you are working with.
\end{enumerate}
\subsection{Conclusion}
In conclusion we have demonstrated a robust repeatable method for topic modelling using the AutoTrader note data. The sector specialist input during the method hyperparameter tuning allows for an expert knowledge sense check of the notes before further analysis and dissemination to a wider team. Feedback from the AutoTrader team noted that the method revealed several technical terms that were previously missed in the topics using their previous expert led technical term search. Additionally, the topic modelling was noted to produce two additional topics that were relevant but not expected to be seen within the content of the notes provided.
\section{Topic modelling utilising Keyword Extraction}
\subsection{Introduction}
The last section looked at topic modelling using LDA. The problem with LDA was that it provided little information as to why the topic was chosen or the subject of the topic. The lack of explanation for the topics is detrimental to understanding them. The first steps of this work are dedicated to find a solution to this problem and apply NLP techniques to help describe the context of identified topics by LDA. The data was cleaned in the same way as the last section but now we had access to live note data which required some more cleaning. For this reason, additional pre-processing steps were added based on a heuristic knowledge obtained from an exploration of the data. These steps were:
\begin{enumerate}
    \item Remove new set of tokens that was identified by data exploration, e.g. `—original message—’
    \item Remove notes shorter than 20 characters assumed to be accidental or not intended to be a note, e.g. `insert text’, `test’, `hjbyhkjbkh’
    \item Remove set of special characters resulting from wrong parsing, e.g. single characters appearing as `\^{a}C\textsuperscript{TM}’
\end{enumerate}
The first step taken to provide more information about the notes, was to work on the topics provided with the use of the LDA analysis. The LDA identified topics within notes and grouped the notes into these topics, but assigned them no description or name. For a user, this means that they are presented with a topic, some group of notes, and they are left to find the meaning of it themselves. This could be enhanced by using the approach of KE on the topics, as keywords should best describe the subject of the text in these topics. Extracting the most important words within the topic should give information about the topic as a whole. Main challenges rising from the AutoTrader dataset for KE are considered to be :

\begin{enumerate}
    \item Dataset obtains large volume of notes, requiring an efficient algorithm that would practically run quickly
    \item Language and jargon of the dataset is unique to the automotive industry, making it even more necessary for the algorithm to be able to encapsulate the subject and meaning notes have
    \item Notes vary in text range, with some of them containing only the important information from a conversation, approach based purely on statistical appearance may not be enough to cope with the lack of text.
\end{enumerate}

We compare RAKE and KeyBERT. The chosen way of evaluating the KE is with feedback from sector experts.
\subsection{Results}
\subsubsection{RAKE and KeyBERT applied to LDA topics}
The first method applied in the project was RAKE, with its potential to quickly and efficiently deliver keywords. LDA was executed again in the same way as the last section, to generate topics. To prepare pre-processed data for RAKE, lemmatised words of each note were joined into single strings. RAKE was run on these strings for each topic to generate keywords for them. To evaluate the keywords, the following action was taken:

\begin{enumerate}
    \item General sanity check of keywords by answering questions:
    \begin{enumerate}
        \item Are keywords short and brief?
        \item Are keywords readable and understandable?
    \end{enumerate}
    \item Generate word cloud graphs for the topics
    \item Compare the words of word clouds with their relative keywords
    \item Together with members of the AutoTrader team, compare word cloud graphs to the extracted keywords and consider their meanings in regards to the automotive industry
\end{enumerate}

For the first point, RAKE extracted extensively long keyphrases, which were too long to be easily readable. For further evaluation to make sense, the length of keywords was limited to maximum of ten words. With now shorter keywords, they were made up of lemmatised words from the pre-processing, forming a set of ten words that together do not create a grammatically consistent sentence. However, some understanding can be achieved by finding correlations between the words of the keyphrases. As an example, in Figure 9, RAKE identified the most important keyphrase containing words `issue upload video go online’, which could translate to the customers having issues with uploading videos online on AutoTrader’s web page. When the word clouds were generated, the graph’s and keyphrases’ words were compared. When comparing each word cloud with its related keyphrases of a topic, it seemed that the keyphrases only sometimes included words contained in word clouds and when they did, normally a single keyphrase included had only one word from the word cloud. When comparing the word cloud to the keyphrase pairs among each other for different topics, they tended to be similar in information, suggesting that topics, and therefore their keyphrases as well, were broad and inconsistent with each other within topic, not bringing any new specific information to the topics. 

\begin{figure}[h]
    \centering
    \includegraphics[width=1\textwidth]{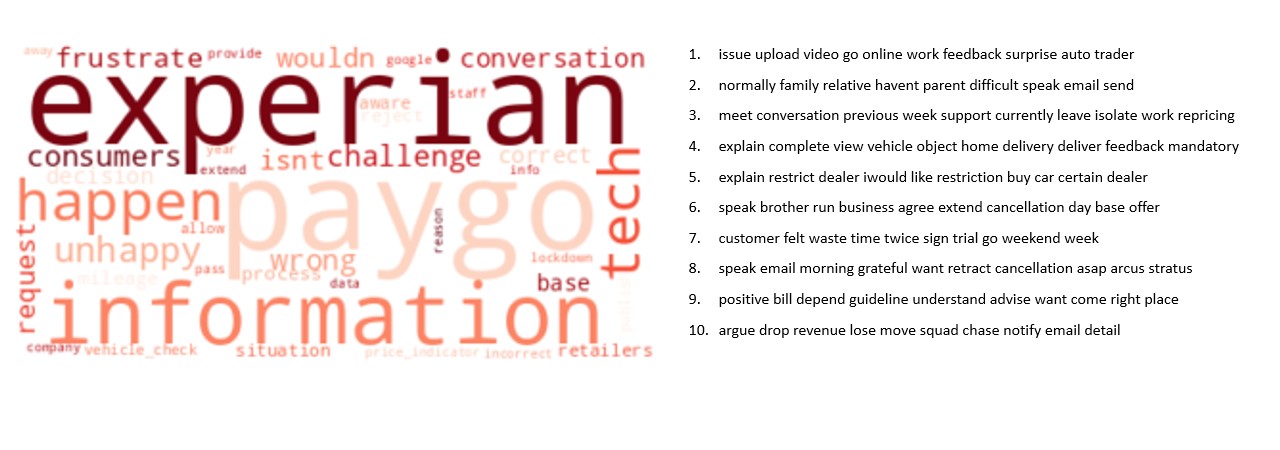}
    \caption{Word cloud - RAKE keyphrases on topic 2.}
    \label{fig:Cloud-RAKE}
\end{figure}

Afterwards, the group of sector experts were presented with a set of word cloud and keyword pairs. An example of this is shown in Figure \ref{fig:Cloud-RAKE} for topic number 2. A set of words may be useful to get a sense of what the subject is about, but it was not practical enough to easily bring insight into the data, leaving major portion of the work to be still done by hand. The foremost issue needed tackling was agreed to be the need of the keywords to be more comprehensive, intelligible and a subject of a topic should be more easily identified from them.

To provide more comprehensible keywords, the next approach was to apply KeyBERT. The idea is that embeddings from BERT should provide deeper understanding of notes, although requiring more computational resources. The implementation of KeyBERT takes as an input one whole text, which meant the preprocessed data had to be formed into a single text. Firstly, sentences were formed by joining words into sentences, separating words with blank spaces. Secondly, these sentences were separated by dots and joined into a single text. All notes that were chosen to be together in a single text, belonged under the same topic group identified by the LDA. Hence, KeyBERT was run on text, that consisted of lemmatised words from notes, separating notes by dots, as if they were sentences, for each of the topics. The parameters of KeyBERT were selected to tackle the issues which arose from RAKE. To have keyphrases contain a lower number of words, the n-gram range was set to be from 2 to 3, expecting keyphrases to have either 2 or 3 words. The `top n’ number of keyphrases was left with the default option on top 5. The keyphrases evaluation was similar to the one of RAKE, having RAKE’s results as a base line:

\begin{enumerate}
    \item General sanity check of keywords by answering questions:
    \begin{enumerate}
        \item Are keywords short and brief?
        \item Are keywords readable and understandable?
        \item How do they compare to RAKE's results?
    \end{enumerate}
    \item Generate word cloud graphs for the topics
    \item Set the words contained in word clouds side by side with their relative keywords and compare results to RAKE
    \item Repeat the previous step with members of the AutoTrader team, weight if the issues were tackled and consider future options
\end{enumerate}

\begin{figure}[h]
    \centering
    \includegraphics[width=1\textwidth]{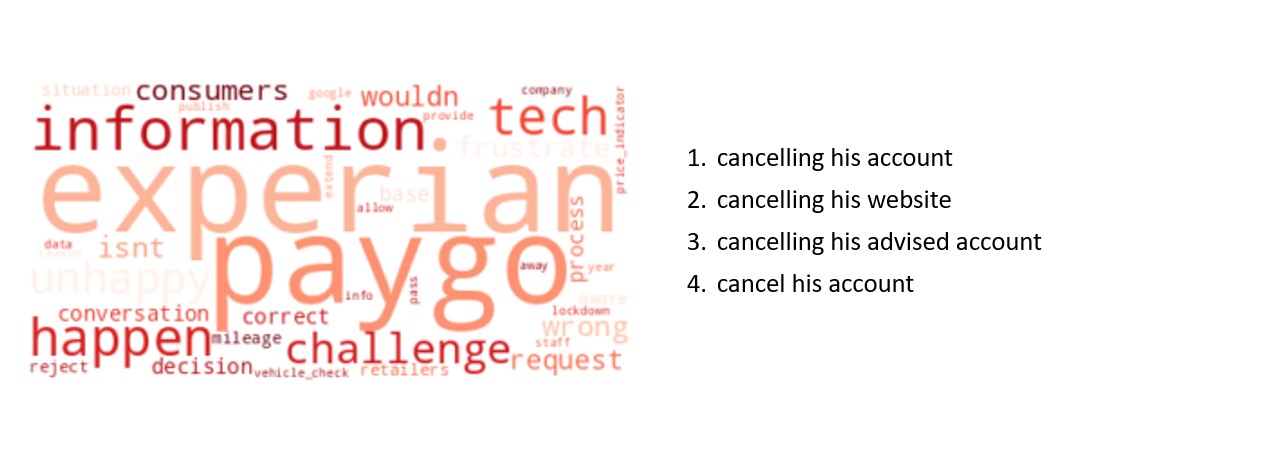}
    \caption{Word cloud - KeyBERT keyphrases on topic 2.}
    \label{fig:Cloud-BERT}
\end{figure}

KeyBERT extracted mostly 3 word n-grams with 2 word n-grams only occurring infrequently. Keyphrases consisted of combinations of words, that together seemed to have a grammatical structure as well as some meaning, e.g. `concerns around priceindicator’. Compared to RAKE, they are more easily interpretable and the information is quickly readable from them. After word clouds were created, there were instances when it was clear that the subject was captured in the keyphrases.

On the other hand, keyphrases appeared to start with the same words within a topic. Furthermore, they seemed to be similar in word choice, sometimes the difference only being the order of the words. This similarity in keywords does not bring any new information about the context of the topic. Compared to RAKE, however, it gives more interpretable and more useful insight into the subject a topic could be about. Following that, the team of experts was presented with the word cloud and keyphrase pairs, in the same way as with RAKE. The expert feedback was that this approach was an improvement on the the RAKE method but the pairs were still too dissimilar.

Issues were noted in both variations of KE algorithms. With the lack of measurements of accuracy for KE algorithms on an unlabelled dataset, word clouds with clear meanings were needed to assess the generated keywords. However, this was not delivered by the LDA algorithm, as words in the word clouds did not always appeared to be related. A new suggestion was made to try to find a new way to categorise notes into topics. From experiments conducted until that point, statistical approaches were seen as insufficient when dealing with given dataset. Inspired by the better performance of BERT, the new approach should capitalise on the semantic approach of BERT-based algorithms to identify topics.
\subsubsection{Rake and KeyBERT applied to topics generated by K-means clustering}
\begin{figure}[h]
    \centering
    \includegraphics[width=1\textwidth]{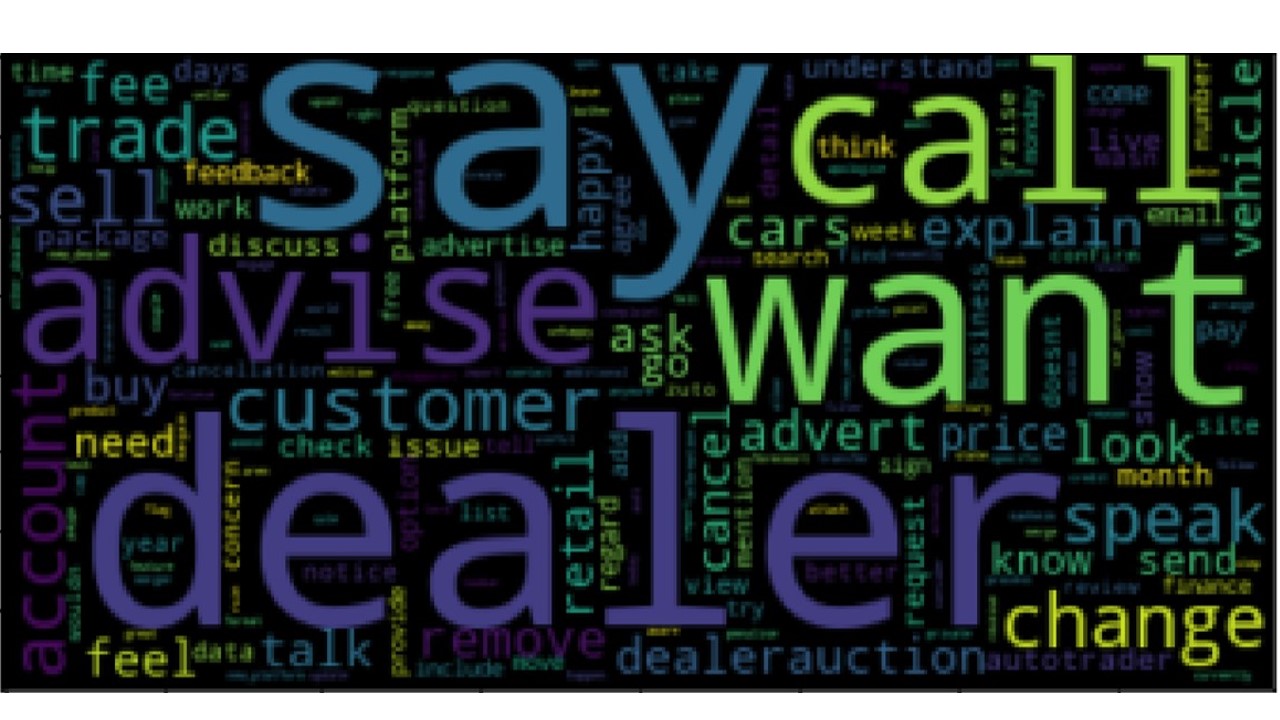}
    \caption{Word cloud using K-Means of cluster 1.}
    \label{fig:K-cluster-cloud}
\end{figure}

The topics provided by LDA were considered as the issue preventing coherent keyword detection, since KE algorithms did not manage to return sane keywords consistently or keywords clearly specific to a topic. This could mean that topics identified by LDA were overlapping or it was not successful at identifying topics. AutoTrader note data is too varied and could be too difficult for the LDA to pick up on the topics and identify them correctly. The solution decided on was to try more modern approaches to Topic Modelling (TM) with methods that can capture the meanings of notes and generate trends based on them. As was in the case of KE, a BERT-based approach was seen to be more successful than statistical methods. Based on that, the next step was to try to get topics using BERT-based embeddings in combination with K-Means clustering. A BERT-based sentence-transformer from HuggingFace library would be first used to encode embeddings for the data. The all-MiniLM-L6-v2 \citep{Hugging2023sentence} transformer model was chosen, because it was designed to map sentences to a 384 dimensional dense vector space and was meant for the application of clustering and semantic search. The embeddings would be then used to cluster them into topics by K-Means clustering method. K-Means is frequently used in NLP clustering and provides easily computable solution to clustering problems. The main issue was the choice of \emph{K}. To find optimal way to determine \emph{K} a rigorous research was conducted. Keeping in mind the interactive goal of the dashboard, the aim was to find the least complicated solution. The most appropriate solution was the ’rule of thumb’, as mentioned in \citet{Naeem2018study}. It is a heuristic approach that does not have a mathematical proof, but is still preferred by researchers. The ’rule of thumb’ formula for \emph{K} is:
\begin{equation}
     K=\sqrt{\frac{n}{2}} 
\end{equation}

where \emph{n} is the number of data points. In our case, \emph{n} is the size of the entire dataset. The aim of this approach was to get results that make more sense than when using LDA method. To compare it, same comparisons with word clouds would be made as when LDA was evaluated. This way a sanity check could be performed to evaluate the success of this approach. This new approach was run on a dataset used by the previous project, containing 10544 notes. The number of topics was computed by the ’rule of thumb’ to be 22. An example of a word cloud from identified generated topic is shown in Figure \ref{fig:K-cluster-cloud}, Both RAKE and KeyBERT were subsequently applied on the generated topics.

In the case of RAKE, keyphrases again contained an abundance of words and the result had to be trimmed down to 10 words per keyphrase to allow us to make judgement. As for the evaluation of this approach, RAKE keyphrases obtained words that were judged to have more meaning by the expert panel as compared to the LDA approach. BERT-based clustering in combination with RAKE seemed to produce more comprehensive keywords. Keyphrases still included a lot of words, resembling the word clouds they were supposed to replace.
\subsection{Recommendations}
\begin{enumerate}
    \item TM algorithms recognise a given number of topics. Although there exist methods to optimise the number of clusters, they are costly in computational resource. This optimisation would have to be run every time new notes are added to the dataset. That happens on a daily basis. So this method does not work well with live data. When TM is run twice on the same data, it generates topics that, although are similar, differ. If TM was run on future datasets, it would create a new set of topics that may not align with previously created topics. This would make topic analysis over a long period of time impossible.
    \item Decide on an agreed objective or milestone before beginning TM analysis. Without a clear end goal the process of selecting new methods and refining existing ones can continue indefinitely. An appropriate objective could be identifying the most prominent keywords associated with the most important topics. The number of these should be agreed at the beginning of the analysis.
\end{enumerate}
\subsection{Conclusion}
KeyBERT again proved to provide concise keyphrases. In contrast to RAKE, KeyBERT’s keyphrases were shorter and neatly formed to contain some meaning. KeyBERT coupled with BERT-based clustering produced the most sane keywords for the topics. It was hard to see, however, if the main goal was achieved. The keywords merely seemed to contain specific words from the word clouds, no new information was obtained.

After the AutoTrader team was shown the word cloud - keyphrases pairs obtained by using sentence transformer with K-Means, it performed a sanity check same as was done previously. The team concluded that although there is stronger correlation between the word cloud with its keyphrases, there is no guarantee that they are actually correlated. Moreover, team stressed the lack of new insights into the data even when more modern approaches were used.
\section{Information retrieval}
\subsection{Introduction}
We then decided on a new approach and the next objective was to change the goal itself. Instead of trying to generate topics, information in the data could be retrieved. Information retrieval (IR) would allow users to search notes and have data visualisation done. Semantic search is a method that provides a NLP solution to the IR problem. It takes a query from the user and enables them to search for notes relating to the given query.
\subsection{Methodology}
During this project, BERT-based approaches had achieved the best results and so it was again chosen to assist with IR. Semantic search could be executed by the use of embeddings from a BERT-based model. Embeddings are high dimensional vectors that should capture the sentimental meaning of a text. The more similar two vectors are, the more similar meanings of two texts are. Vector similarity is attained with the use of cosine similarity function. The main idea of this approach is to accomplish this task by executing these steps:

\begin{enumerate}
    \item Use BERT-based model to encode embeddings for all notes
    \item Use the same BERT-based model to create an embedding for a given query
    \item Compute cosine similarity between the query embedding and all note embeddings, this is named a similarity score, and assign embeddings to their respective notes
    \item Compile a list of notes based on ordering notes by their similarity from highest to lowest.
\end{enumerate}

In the previous approaches in this project, only qualitative analysis could be performed without manually labelling thousands of notes. So to obtain a quantitative analysis the labelling was required. Manually labelling notes can take time and the input of sector experts is of great assistance in this task. For the labelling the following approach was designed:

 \begin{enumerate}
     \item With the help of the AutoTrader team, agree on 7 topics that could be queried
     \item Create a labelling dataset by randomly selecting 1000 notes
     \item Manually label the dataset by noting 1/0 as to whether the note belongs to the topic, consulting the team to try to minimalise bias
     \item Use a ranking score to assess the ranking accuracy of this approach
 \end{enumerate}

For this task, the labelling dataset was constructed by randomly choosing 1000 notes from the live dataset from year 2021. Admittedly, it was not a large dataset, but it would still help to create a certain sense of how this approach is behaving. Two annotators worked to label the dataset and one annotated 169 notes and the other annotated all 1000. Agreement between the annotators was calculated using Cohen's kappa \citep{Cohen1960coefficient}. The result was 0.64 indicating substantial agreement. Topics agreed upon were:

\begin{enumerate}
    \item Valuation
    \begin{enumerate}
        \item Note contains feedback or valuation on AutoTrader’s services
        \item Considering that this AutoTrader dataset should consist mainly of feedback, it is used to distinguish notes that do not contain feedback
        \item Example text: \emph{feels like service is X years out of date}
    \end{enumerate}
    \item Price
    \begin{enumerate}
        \item Note discusses a price of services or products, when in context of the AutoTrader’s prices or the dealer’s prices
        \item Example text: \emph{advanced to help with this by increasing his prices, PRODUCTNAME he loves but didn’t realise it was MONEY per month}
    \end{enumerate}
    \item Package
    \begin{enumerate}
        \item Note discussing AutoTrader product package
        \item Example text: \emph{PRODUCT-NAME not performing as well it he hoped to, PRODUCT-NAME he loves but didn’t realise it was MONEY per month}
    \end{enumerate}
    \item Cancellation
    \begin{enumerate}
        \item Note considers cancelling a service or informs about cancelling it
        \item Example text: \emph{process the cancellation to downgrade}
    \end{enumerate}
    \item Stock
    \begin{enumerate}
        \item Stock in AutoTrader’s journal refers to a vehicle being sold on their website
        \item Note talks about some stock
        \item Example text: \emph{uses all auction sites to sell it as well}
    \end{enumerate}
    \item Tech
    \begin{enumerate}
        \item Note mentions any of AutoTrader’s online services
        \item Example text: \emph{gets an error when you click on ’see’, allows the upload for images only}
    \end{enumerate}
    \item Billing
    \begin{enumerate}
        \item Note contains information about money transport or concerns about it
        \item Example text: \emph{at the moment not selling well but has to pay X outstanding money}
    \end{enumerate}
\end{enumerate}

As for the ranking score, a Normalised Discounted Cumulative Gain (NDCG) was chosen. NDCG provides a relatively easy to compute ranking score. It is deemed to be a ”classic” information retrieval (IR) metric. The goal of NDCG is to rank results from IR by comparing them to an ideal rank ordering. NDCG is computed from the discounted cumulative gain DCG as defined in \citet{Agrawal2009diversifying}.The process of evaluation was as follows:
\begin{enumerate}
    \item Let us have a query Q belonging to one of the topics
    \item Query the note corpus with Q
    \item Provide NDCG scores for all topics
\end{enumerate}

The idea behind this process is that if the query Q belongs to a topic, NDCG score for this topic should be higher. If the query Q does not belong to a topic, it should drop. As the querying process returns an ordered list of notes with similarity scores, similarity scores can be compared to 1s and 0s from the labelling dataset to achieve NDCG score. This way, the NDCG score compares the created similarity scores to an ideal ranking. The result of this are NDCG scores, which can be compared to a baseline. This baseline is that scores for all notes are 0.5. Another analysis could be performed to see the impact of pre-processing on the NDCG scores. To see the impact of this, the same NDCG analysis is performed on pre-processed notes, as well as original notes. Computed NDCG scores for them are shown in table \ref{table:2} and table \ref{table:3}.

\begin{table}[h]
\centering
\begin{tabular}{|c|c|}
\hline
Clean & Baseline\\
\hline
Valuation &0.97 \\
\hline
Price &0.76 \\
\hline
Package &0.78 \\
\hline
Cancellation &0.63 \\
\hline
Stock &0.68 \\
\hline
Tech &0.86 \\
\hline
Billing &0.56 \\
 \hline
\end{tabular}
\caption{ Baseline NDCG evaluations for the clean data.
}
\label{table:2}
\end{table}

\begin{table}[h]
\centering
\begin{tabular}{|c|c|}
\hline
Pre-processed & Baseline\\
\hline
Valuation &0.97 \\
\hline
Price &0.76 \\
\hline
Package &0.78 \\
\hline
Cancellation &0.63 \\
\hline
Stock &0.68 \\
\hline
Tech &0.86 \\
\hline
Billing &0.57 \\
 \hline
\end{tabular}
\caption{ Baseline NDCG evaluations for the pre-processed data.
}
\label{table:3}
\end{table}

From observing the Baseline it is clear, that topics are not equally represented in the labelled dataset. Because all values are 0.5, higher NDCG for valuation means that there is higher volume of 0.5 data points. There is not difference between clean and pre-processed data.

HuggingFace provides many models for the task of semantic search. A set of them was picked to be analysed. The 768 dimensional vector models on which an analysis was performed were compared and the multi-qa-mpnet-base-dotv1 \citep{Hugging2023transformers} model was chosen to be implemented. Only the pre-processed notes are examined moving forward.
\subsection{Results}
Across all models the valuation scored highly, similar to the baseline models this is due to the notes all containing the feedback flag from the AutoTrader team. The model was further tested with the use of queries. These queries were designed to test the model and provided a use case for how the queries could be used in the future. The first query was `tech issue'. This was expected to result in a higher than baseline score for the tech category. Table \ref{table:4} shows the scores for each topic given this query:

\begin{table}[h!]
\centering
\begin{tabular}{|c|c|c|}
\hline
Topic & Score & Difference from baseline\\
\hline
Valuation &0.96 & -0.01\\
\hline
Price &0.72 & -0.04\\
\hline
Package &0.74 & -0.04\\
\hline
Cancellation &0.60 & -0.03\\
\hline
Stock &0.67 & -0.01\\
\hline
Tech &0.92 & +0.06\\
\hline
Billing &0.55 & -0.02\\
 \hline
\end{tabular}
\caption{NDCG evaluations for the query ``tech issue" on the pre-processed data.}
\label{table:4}
\end{table}

As expected the query resulted in a higher score for tech. All other scores decreased showing that the query was less relevant to those topics. The next query given was ``too expensive". This was anticipated to increase the relevance of the price topic. The result is given below in table \ref{table:5}:

\begin{table}[h!]
\centering
\begin{tabular}{|c|c|c|}
\hline
Topic & Score & Difference from baseline\\
\hline
Valuation &0.97 & 0.00\\
\hline
Price &0.86 & +0.10\\
\hline
Package &0.79 & +0.01\\
\hline
Cancellation &0.66 & +0.03\\
\hline
Stock &0.68 & 0.00\\
\hline
Tech &0.82 & -0.04\\
\hline
Billing &0.54 & -0.03\\
 \hline
\end{tabular}
\caption{NDCG evaluations for the query ``too expensive" on the pre-processed data.}
\label{table:5}
\end{table}
These values show an increased score for price as expected but also have higher then baseline values for package and cancellation. This could be interpreted that packages that are too expensive can lead to cancellations. This is a logic assessment of how AutoTrader's customers show their feedback to high prices. This method of inputting queries can demonstrate the relevant topics to the inquirer. The last query tested was "send money", this query should highlight the billing topic and also may highlight the price topic.

\begin{table}[h]
\centering
\begin{tabular}{|c|c|c|}
\hline
Topic & Score & Difference from baseline\\
\hline
Valuation &0.96 & -0.01\\
\hline
Price &0.75 & -0.01\\
\hline
Package &0.74 & -0.04\\
\hline
Cancellation &0.65 & -0.02\\
\hline
Stock &0.64 & -0.04\\
\hline
Tech &0.84 & -0.02\\
\hline
Billing &0.68 & +0.11\\
 \hline
\end{tabular}
\caption{NDCG evaluations for the query ``send money" on the pre-processed data.}
\label{table:6}
\end{table}
The result here shows that the billing topic does increase in relevance significantly compared to the previous queries and the baseline. The other topics are at values decrease in relevance compared to the baseline showing that the billing topic may have a distinct set of notes that are only pertinent to the paying of bills.

The current approach enables the user to order the notes based on the similarity to a given query. This list includes all notes. Should the user be presented with analysis based on given query, a subset of notes has to be created from this list. The subset would include notes that are more similar to the query than notes that are not included in the subset. Graphical analysis would then be done on notes from the subset. 
\subsection{Recommendations}
We also have recommendations for data retrieval. 
\begin{enumerate}
    \item Get a number of sector experts to annotate if possible and compare the annotations, then reject annotators if they vary too much from the standard. This method increases the reliability of annotators and removes the possibility of one annotator from being indistinguishable from the ground truth. If they are sector experts then they will better understand the context behind the conversations.
    \item Use a variety of models and then select the best performer. For this paper we examined five different models before selecting the best performer. If only one method is used then the performance of the analysis could be worse off than if more were first evaluated.
    \item Try to avoid having a large overlap between evaluation topics. This can lead to situations where topics such as billing are almost a subset of the price topic. Splitting the set of notes into separate categories is easier with diverse topics. Topics should also be limited in scope and if too many notes are within one topic then the topic definition needs to be more narrowly defined.
\end{enumerate}
\subsection{Conclusion}
Information retrieval using topics decided by experts has provided better results than with topics derived from topic modelling. The manually labelled data allowed for a verification of the similarity metric. The queries allow the investigator to find notes similar to the areas of investigation. Unlike the topic modelling work the results returned here could be replicated. The analysis also is fast enough to be updated each day for the relevant data. The similarity threshold allows for a relevant number of notes to be analysed.
\section{Conclusion and future work}
This paper has looked at the analysis of business to business data with natural language processing. Sentiment analysis, topic modelling and information retrieval are applied to the data. The sentiment analysis work allowed the data to be analysed automatically and can help AutoTrader by reducing the time spent on this task. The approach is transferable and could be used to automatically analyse other note data in different areas. The topic modelling did not manage to achieve the results desired and lacked clarity. Informational retrieval worked more effectively.

The work struggled with the unlabelled nature of the notes and the masking of words in the sentences further removed information that would have been helpful for analysis. The annotated data could be expanded upon with more annotators and a larger number of notes studied. Recently, Large Language Models (LLMs) have an increased profile in the area of NLP \citep{Qin2023chatgpt}. Further work could be done to train a LLM on the notes studied here. This model could then be prompted to answer questions about topics within the notes.




\end{document}